\documentclass[10pt,twocolumn,letterpaper]{article}

\usepackage[pagenumbers]{cvpr} 

\usepackage{graphicx}
\usepackage{appendix}
\usepackage{amsmath}
\usepackage{amssymb}
\usepackage{booktabs}
\usepackage{multirow}
\usepackage{bm}
\usepackage{bbm}
\usepackage{color}

\usepackage[capitalize]{cleveref}
\crefname{section}{Sec.}{Secs.}
\Crefname{section}{Section}{Sections}
\Crefname{table}{Table}{Tables}
\crefname{table}{Tab.}{Tabs.}

\def\cvprPaperID{4297} 
\def\confName{CVPR}
\def\confYear{2023}

\begin{document}

\title{Glocal Energy-based Learning for Few-Shot Open-Set Recognition}

\author{
Haoyu Wang\textsuperscript{1}\thanks{H. Wang, G. Pang and P. Wang contributed equally in this work.} \quad
Guansong Pang\textsuperscript{2}\footnotemark[1] \quad
Peng Wang\textsuperscript{3}\footnotemark[1] \quad
Lei Zhang\textsuperscript{1} \quad
Wei Wei\textsuperscript{1} \quad
Yanning Zhang\textsuperscript{1}\thanks{Corresponding author.} \\
\textsuperscript{1}Northwestern Polytechnical University \\
\textsuperscript{2}Singapore Management University \qquad
\textsuperscript{3}University of Wollonong\\
}

\maketitle

\begin{abstract}

  Few-shot open-set recognition (FSOR) is a challenging task of great practical value. It aims to categorize a sample to one of the pre-defined, closed-set classes 
   illustrated by few examples while being able to reject the sample from unknown classes. In this work, we approach the FSOR task by proposing a novel energy-based hybrid model. The model is composed of two branches, where a classification branch learns a metric to classify a sample to one of closed-set classes and the energy branch explicitly estimates the open-set probability. To achieve holistic detection of open-set samples, our model leverages both class-wise and pixel-wise features to learn a glocal energy-based score, in which a global energy score is learned using the class-wise features, while a local energy score is learned using the pixel-wise features. The model is enforced to assign large energy scores to samples that are deviated from the few-shot examples in either the class-wise features or the pixel-wise features, and to assign small energy scores otherwise. Experiments on three standard FSOR datasets show the superior performance of our model.\footnote{Code is available at \url{https://github.com/00why00/Glocal}}

\end{abstract}

\section{Introduction}

In recent years, deep learning has flourished in various fields
with the ever-increasing scale of the training data under the closed-world learning settings,
\ie, the training and test sets share exactly the same set of classes. However, such settings often do not hold in many real applications.
This is because 1)
it is difficult or costly to obtain a large amount of labeled data, and 2) models deployed in open-world environments need to constantly deal with samples from unknown classes. For example, in the application of deep learning for diagnosing rare diseases, the number of samples is limited. In this case, the model is prone to over-fitting, resulting in a significant degradation in performance. Further, there can be unknown variants of those diseases due to our limited understanding of the diseases.
Thus, the models are required to perform the classification accurately for the classes illustrated by limited samples, while at the same time detecting the samples from unknown classes. The latter ability is important, especially for healthcare or safety-critical applications, \eg, to alert the unknown cases for human investigation in the disease diagnosis example, or to request human intervention for handling unknown objects
in autonomous driving.

\begin{figure}[t]
  \centering
  \includegraphics[width=0.95\linewidth]{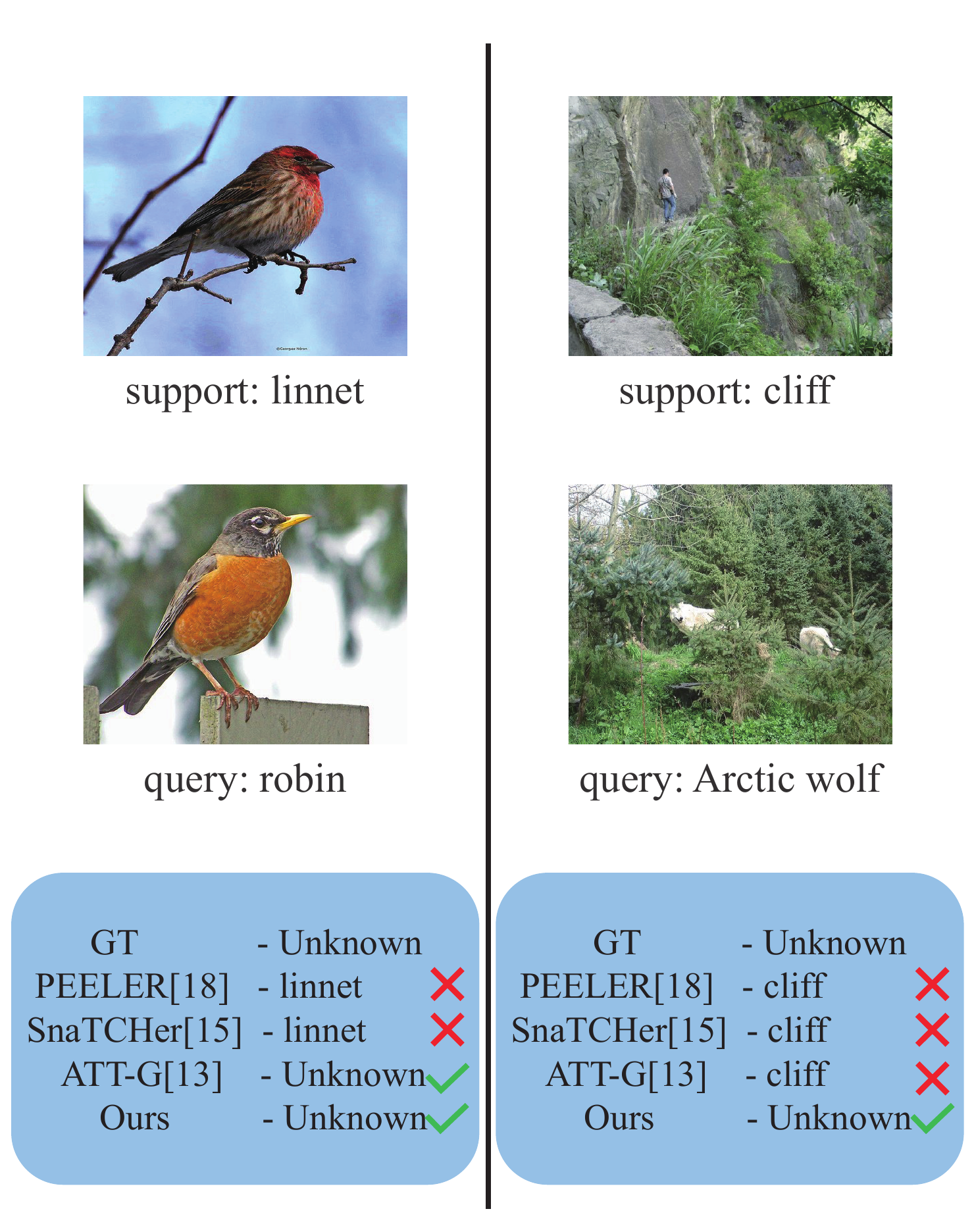}
  \caption{Two typical errors with existing methods. Our method can detect unknown open-set samples that are similar to the closed-set samples in either the global class level or the local feature level.
  }
  \label{fig:pixel motivation}
\end{figure}

Few shot learning~\cite{ProtoNet, Matching, Siamese, O1, M1, MA1} (FSL) and open set recognition~\cite{OT1, OT2, D1, D2, G1, G2} (OSR) are two techniques dedicated to solve these two problems, respectively. FSL methods are trained to achieve a good generalization ability on the new task with only a few training samples. 
But FSL approaches 
are developed under a closed-set setting. It lacks the ability to distinguish the classes unseen during training.
The goal of OSR is, on the other hand, to recognize open-set samples while maintaining the classification ability of closed-set samples. 
However, its classification ability is often built upon the availability of a large number of training samples. Thus, OSR approaches fail to work effectively when only a few training samples are available.

Few-shot open-set recognition (FSOR), which combines the FSL and OSR problems, is a 
largely under-explored area. FSOR requires the model to utilize only a few training samples to effectively achieve the ability of both closed-set classification and open-set recognition. 
Existing FSOR methods~\cite{SnaTCHer,ATT,RFDNet} are based on the prototype network~\cite{ProtoNet}, which performs classification by measuring the distance between the prototype of each class and the query embedding feature of a sample. They improve the original closed-set classifier in recognizing open-set samples by 
learning an additional open-set class using pseudo open-set samples. However, only the class-wise information of the sample is considered, and the pixel-wise spatial information of the sample is ignored. 
As shown in Figure \ref{fig:pixel motivation},
these methods fail to distinguish open-set samples from closed-set class samples that share similar global semantic appearances. 
Further, the optimization objectives of FSL and OSR are different from each other.
Thus, training using only an open-set classifier can limit the performance of these models.

To solve the above problems, we propose a novel FSOR method, called Glocal Energy-based Learning (GEL). 
Different from previous methods, GEL consists of two classification components: one for closed-set classification and one for open-set recognition. 
Specifically, in addition to use the class-wise features to classify closed-set samples in the closed-set classifier, GEL leverages both class-wise and pixel-wise features to learn a new energy-based open-set classifier, in which a global energy score is learned using the class-wise features while a local energy score is learned using the pixel-wise features. GEL is enforced to assign large energy scores to samples that are deviated from the few-shot examples in either the class-wise features or the pixel-wise features, and to assign small energy scores otherwise. In doing so, GEL can detect unknown class samples that are deviated from the known classes in either high-level abstractions or fine-grained appearances, as shown in Figure \ref{fig:pixel motivation}. 

In summary, this work makes the following three main contributions:
\begin{itemize}
  \item We propose a novel FSOR framework that 
  learns glocal 
  open scores for detecting unknown samples from the class-wise (global) and pixel-wise (local) scales.
  \item We 
  further propose a novel energy-based FSOR model, dubbed GEL, that learns glocal energy-based open scores based on the class-wise and pixel-wise similarities of query samples to the support set.
  \item Through extensive experiments on three widely-used datasets, we show that GEL outperforms state-of-the-art competing methods and achieves state-of-the-art results on these benchmarks.
\end{itemize}

\section{Related Work}

\subsection{Few-Shot Learning}

Few-shot learning has been widely studied in computer vision. The approaches of FSL can be divided into two categories: meta-learning based approaches and transfer learning approaches. There are three subcategory of meta-learning based approaches. The first is metric-based approaches~\cite{ProtoNet, Matching, Siamese} which learn a distance function through training samples. Another subcategory is optimization-based approaches~\cite{O1, O2, O3, M1, M2, M3, M4, M5}, it learns a priors to optimize the model on limited training examples without overfitting. The last subcategory is model-based approaches~\cite{MA1, MA2, MA3, MA4, MA5}. Different from the previous two approaches, model-based approaches use support samples to generate model weights adapted to new task. Another category of FSL is transfer learning~\cite{T1, T2, T3}. Transfer learning improve the performance in a new task through transferring knowledge from a learned related task.

\subsection{Open-Set Recognition}

Open-set recognition is a more realistic scenario because it is usually difficult to include all classes when training a classifier. OSR requires the classifier to classify not only known classes, but also unknown classes. There are two mainstream approaches for OSR. One is discriminative model. It includes traditional machine learning methods~\cite{OT1, OT2, OT3, OT4, OT5} and deep neural network methods~\cite{D1, D2, D3, D4}. The former is popular before deep neural network rise. It usually adapt the limitation of traditional methods that training and testing data are from the same distribution for OSR. The latter uses the powerful representation ability of deep neural network to solve OSR through network architecture design. The other mainstream approach is generative model~\cite{G1, G2, G3, G4}. It usually use generative adversarial network or Dirichlet Process to generate unknown samples as training samples to improve model performance.

\begin{figure*}[t]
  \centering
  \scalebox{0.9}{
  \includegraphics[width=1.0\linewidth]{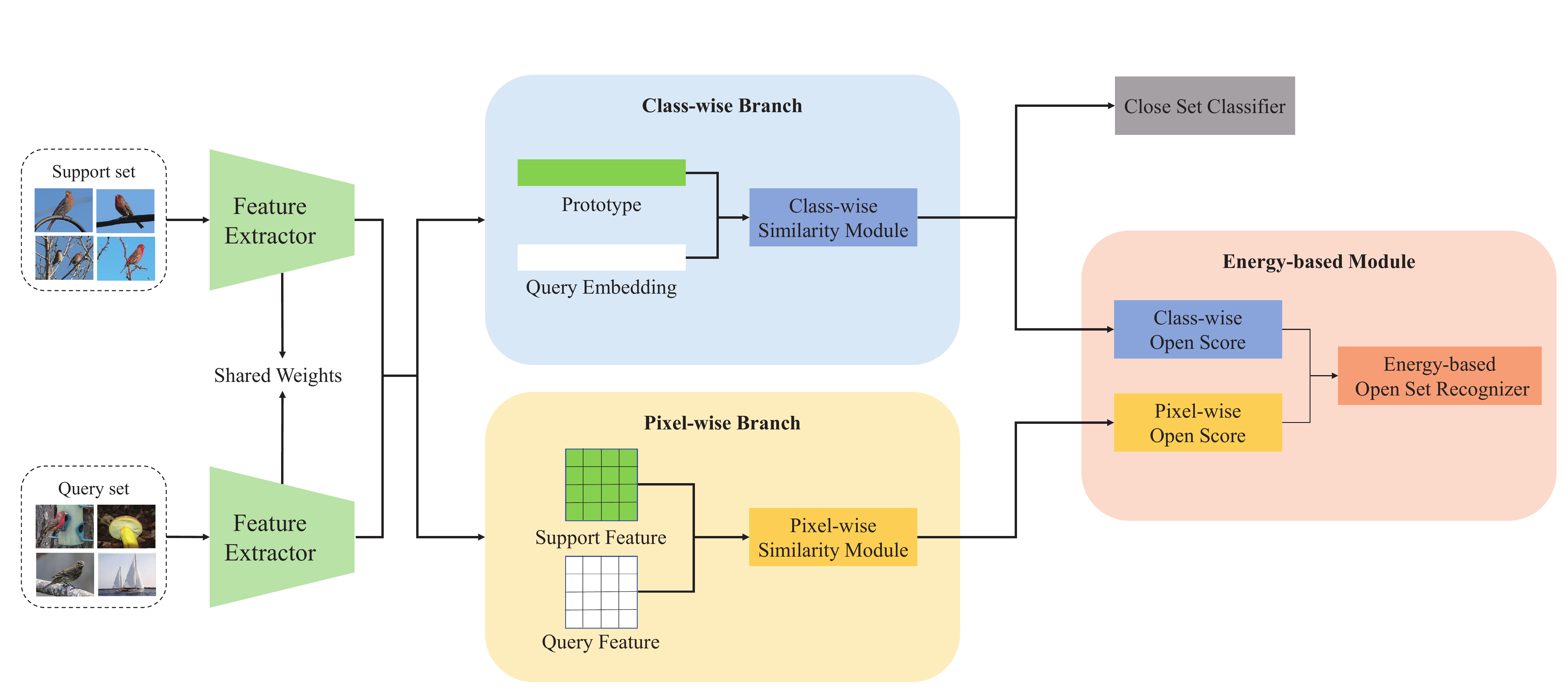}}
  \caption{An overview of our model GEL. Our model mainly consists of three parts: a \textbf{class-wise branch}, which is used to calculate the similarity between embeddings for closed-set classification and open-set recognition, a \textbf{pixel-wise branch}, which is used to calculate the similarity between feature maps for open-set recognition, and an \textbf{energy-based module}, which performs open-set recognition on query samples by feeding the similarities from the two branches to learn glocal energy scores.}
  \label{fig:overview}
\end{figure*}

\subsection{Few-Shot Open-Set Recognition}

Compared with FSL and OSR, there are few studies focused on few shot open-set recognition, which mainly include the following three methods. The first is the loss function based method. Based on the original distance-based classifier, PEELER~\cite{PEELER} proposes an open-set loss to improve the accuracy of open-set recognition by increasing the entropy of classification results of open-set samples. The second method is based on transformation consistency. SnaTCHer~\cite{SnaTCHer} adds each query embedding replacement to the prototype set, and determines whether the query sample is an open-set sample by measuring the difference between the sets before and after transformation. By measuring the difference before and after the change of the set, the unknown class distribution estimation problem is transformed into a relative feature transformation problem which is unrelated to the unknown class samples. The third method is to add an extra open-set class. TANE~\cite{ATT} and RFDNet~\cite{RFDNet} expands the closed-set classifier by using a generative network to obtain an additional open-set classes prototype from closed-set prototype. They then add the prototype to the original classifier to enable it to perform both closed-set classification and open-set recognition. By adding an additional category to classify the query sample, the measuring based on the entropy or the threshold of the sample is changed to dynamic classification.

\section{Preliminary}

The goal of the FSOR is to identify open-set samples while maintaining the classification capability of closed-set with only a few data samples. In particular, for a N-way K-shot Q-query sampled from dataset $ \mathcal{D} $, the FSOR task can be presented as: $ \mathcal{T} = \{\mathcal{S}, \mathcal{Q}_{k}, \mathcal{Q}_{u}\} $, where $ \mathcal{S} = \{x_i, y_i\}_{i=1}^{|\mathcal{S}|} $ and $ \mathcal{Q}_{k} = \{x_i, y_i\}_{i=1}^{|\mathcal{Q}_k|} $ are support set and known query set, respectively. The label $ y $ corresponding to the image $ x $ in these two sets are from closed-set categories $ \mathcal{C}_k $. Different from FSL, $ \mathcal{Q}_{u} = \{x_i, y_i\}_{i=1}^{|\mathcal{Q}_u|} $ is a set of query samples from unknown classes
$ \mathcal{C}_u $, and $ \mathcal{C}_k \cap \mathcal{C}_u = \varnothing $.

\section{Method}

We will describe our approach in detail in this section. First, we will give an overview of our proposed method GEL, then introduce our proposed pixel-wise similarity module, and finally introduce our energy-based open-set recognizer.

\subsection{Overview}

Figure \ref{fig:overview} shows the overall architecture of our model. Following previous FSOR methods~\cite{PEELER,SnaTCHer,ATT,RFDNet}, we use metric-based meta-learning architecture. First, the embedding and feature maps of the support and query samples are obtained through a shared feature extractor $ \mathcal{F}_\theta $, where $ \theta $ denots the shared parameters. For each input support sample $ x_i^{\mathcal{S}} $, we obtain its embedding $ e_i^{\mathcal{S}} $ and feature map $ f_i^{\mathcal{S}} $ by $ e_i^{\mathcal{S}} = avg\ pooling(\mathcal{F}_\theta(x_i^{\mathcal{S}})) $ and $ f_i^{\mathcal{S}} = \mathcal{F}_\theta(x_i^{\mathcal{S}}) $, respectively. For each query sample $ x_i^{\mathcal{Q}} $, we use the same way to obtain its query embedding $ e_i^{\mathcal{Q}} $ and feature map $ f_i^{\mathcal{Q}} $. The channel dimension of them is $ dim $.

For the class-wise branch, we first compute the prototype $ p $ for each of the N classes by averaging over support embeddings. Particularly, for a class $ n $, its prototype $ p_n $ is calculated from all support embedding of class $ n $ by

\begin{equation}
p_n = \frac{1}{K}\sum_{i=1}^K\mathcal{F}_\theta(e_{n, i}^\mathcal{S}).
\end{equation}

Then we follow~\cite{SnaTCHer} to simply use a self-attention module to enhance the class prototypes. For the matrix $ P \in \mathbbm{R}^{N \times dim}$ composed by all prototypes, the enhanced prototype matrix $ P^{*} $ is calculated as follows

\begin{equation}
P^q = P W^q,\ P^k = P W^k,\ P^v = P W^v,
\end{equation}

\begin{equation}
P^{*} = softmax(\frac{P^q {P^k}^T}{\sqrt{d_{P^k}}})P^v,
\end{equation}
where $ W^q, W^k, W^v \in \mathbbm{R}^{dim \times dim} $ are coefficient matrices that linearly map the prototype matrix, and $ d_{P^k} $ is the channel dimension of $ P^k $.

Finally, we obtain the class-wise similarity $ s_c $ by measuring the distance between the query embedding and the enhanced class prototype:

\begin{equation}
s_c^{i,n} = -distance(e_i^{\mathcal{Q}}, p_n^{*}),
\end{equation}
where $ s_c^{i,n} $ is the class-wise similarity between $ x_i^{\mathcal{Q}} $ and the enhanced prototype of class $ n $, $ distance(\cdot,\cdot) $ is a distance function. We use Euclidean distance by default, because we found it performed best through experiments.

In closed-set classification, only the class-wise similarity $ s_c $ is used to classify query samples in $ \mathcal{Q}_{k} $ through a softmax function:

\begin{equation}
p(y=n|x_i) = \frac{e^{s_c^{i,n}}}{\sum_{j \in \mathcal{C}_k} e^{s_c^{i,j}}}.
\end{equation}
Then we use cross entropy loss to optimize the model:
\begin{equation}
L_c = \frac{1}{NQ}\sum_{i=1}^{NQ}\sum_{j \in \mathcal{C}_k}\mathbbm{1}_{x_i=j}log(p(y=n|x_i)),
\end{equation}
where $ \mathbbm{1}_{condition} $ is an indicative function, which is one if the condition is met, and zero otherwise, and$ p(y=n|x_i) $ is the probability that the label of sample $ x_i $ is class $ n $.

Our model extends this popular FSL approach to FSOR by using both class-wise similarity and pixel-wise similarity to learn a global energy-based open-set recognizer. We will detail our proposed module in the next two subsections.

\subsection{Pixel-wise Similarity Module}

In order to have a holistic recognition of open-set samples, 
in addition to the class-wise information, we also consider the pixel-wise information. As shown in Figure \ref{fig:pixel motivation}, our idea is based on the following two key points. First, if the class of open-set samples is similar to that of closed-set samples, the distance between their embeddings would be so small, so it is difficult to distinguish them. However, human beings usually distinguish them by a certain key part. Secondly, if the large background of the open-set and closed-set samples are similar, it is often difficult to extract the class information only by using the embeddings. But humans can pick up the key parts directly from the background. Inspired by these two observations, we propose a novel pixel-wise similarity module. First, we use the same method as the class-wise branch to obtain the feature map for each class. Formally, for a class $ n $, its feature map $ f_n $ is defined as

\begin{equation}
f_n = \frac{\sum_{i=1}^{K} f_i}{K}.
\end{equation}

In order to keep the distance between pixel-wise features and class-wise features within the same scale convenient for fusion and reduce the amount of computation, we first apply a scale calibration module to all feature maps. Specifically, we use a point-wise convolution to halve the channel dimension of the feature map, followed by a Batch Normalization layer and a PReLU activation function:

\begin{equation}
f^{*} = PReLU(BN(Conv(f))).
\end{equation}

Finally, for each pixel of the class feature map and query, we calculate the pixel-level similarity using cosine similarity. And for each query pixel, we consider the top-k nearest neighbours in class feature map pixels and calculate the summation for the top-k similarity scores to form a robust fine-grained metric for open-set learning:

\begin{equation}
s_f^{i,n} = \sum_{pixel}topk(\frac{f_i^{*} \cdot f_n}{||f_i^{*}||\cdot ||f_n||}) / T,
\label{eq: temp}
\end{equation}
where $ f_n \in \mathbbm{R}^{\frac{dim}{2} \times m^2}$ and $f_i^{*} \in \mathbbm{R}^{m^2 \times \frac{dim}{2}} $ are the feature map of class $ n $ and query sample $ x_i $, respectively; $ m $ is the spatial dimension of feature map; $ ||\cdot|| $ is applied to the fourth dimension; $ \sum_{pixel} $ sum over all the remaining pixels after calculating $ topk $; and $ T $ is a temperature coefficient.

\textbf{Remark.} Attention mechanism is also a common method to focus on local information in the deep learning approach, but in the few-shot open-set recognition task, this can easily result in over-fitting because of the lack of samples. Our proposed pixel-wise similarity module has only a few learnable parameters in the scale calibration module, well alleviating the over-fitting issue.

\subsection{Energy-based Module}

The existing FSOR methods often use cross-entropy-based classifiers, because for closed-set classification, entropy-based classifiers can often achieve better results by classifying the samples using prediction probability. However, for our dual-branch architecture, we have a dedicated energy-based open-set recognizer to classify open-set samples. 
By using energy-based models, we eliminate the complexity in the process of normalization the probability distribution in entropy-based classification. Thus, they are more suitable for the FSOR task.

Our energy-based module integrates the class-wise and pixel-wise similarity results from the two branches and then feed them to train into glocal energy-based few-shot open-set recognizer.

\subsubsection{Margin-based Energy Loss}

We then feed these glocal energy scores to an energy loss function to optimize our model.
The model is trained to ensure that the energy is large when the samples are from unknown classes, and it is small otherwise. To achieve this goal, we use a  margin-based energy loss to optimize the model. Particularly, for a query sample $ x_i $, its energy loss is calculated by

\begin{equation}
L_e^{x_i} = \left\{
\begin{array}{cl}
max(0, E^{x_i} - M_{k}) &  \mathbbm{1}_{y_i \in \mathcal{C}_k} \\
max(E^{x_i} - M_{u}, 0) &  \mathbbm{1}_{y_i \in \mathcal{C}_u}, \\
\end{array} \right.
\label{eq: margin}
\end{equation}
where $  M_{k} $ and $ M_{u} $ are the margins of closed-set and open-set samples, respectively. $\mathcal{C}_u$ are pseudo open set samples, consisting of $ Q $ samples of each class from $ N $ classes that do not overlap with the closed-set classes $ \mathcal{C}_k $. Both $\mathcal{C}_u$ and $ \mathcal{C}_k $ are part of the training set. The energy loss for the task is

\begin{equation}
L_e = \sum_{\mathcal{Q}_{k} \cup \mathcal{Q}_{u} }L_e^{x_i}.
\end{equation}

Thus, the total loss of our GEL model is 

\begin{equation}
L = L_c + \lambda L_e,
\label{eq: lambda}
\end{equation}
where $ \lambda $ is a hyperparameter that balances the two losses.

\subsubsection{Glocal Energy Function}

For a query sample $ x_i $, its class-wise (global) similarity to class $ n $ is $ s_c^{i,n} $, while its pixel-wise (local) similarity is $ s_f^{i,n} $. We first use these two similarities to respectively define the global and local energy through the energy function as follows:

\begin{equation}
E_c = -log \sum_{n \in \mathcal{C}_k}(e^{s_c^{i, n}}),
\end{equation}

\begin{equation}
E_f = -log \sum_{n \in \mathcal{C}_k}(e^{s_f^{i, n}}).
\end{equation}

Since we have used the scale calibration module in the pixel-wise similarity module to adjust the scale of similarity score, the final glocal open-set energy of the sample can be obtained by a simple addition: 

\begin{equation}
E = E_c + E_f.
\label{eq: combine}
\end{equation}

Other combination methods are evaluated and compared in our ablation study in Sec. \ref{sec ablation}.


\section{Experiments}

\subsection{Datasets}

We use miniImageNet~\cite{miniImageNet}, tieredImageNet~\cite{tieredImageNet} and CIFAR-FS~\cite{CIFSR-FS} to evaluate the performance of the model. MiniImageNet contains a total of 60,000 images of size $ 84 \times 84 $ in 100 categories, including 600 samples in each category. The category for training, validation and testing set is 64, 16, and 20, respectively. TieredImageNet is a larger dataset. It contains a total of 779,165 images of size $ 84 \times 84 $ in 608 categories, including 351 for training, 97 for validation and 160 for testing. Both of them are the subsets of ILSVRC-12~\cite{ImageNet}. CIFAR-FS is the subsets of CIFAR100~\cite{CIFAR100}. It contains a total of 60,000 images of size $ 32 \times 32 $ in 100 categories. The categories are divided in the same way as miniImageNet.

\subsection{Metrics}

To measure the effectiveness of the FSOR methods, following ~\cite{PEELER,SnaTCHer,ATT,RFDNet}, we use ACC and AUROC as the metrics. ACC is used to measure the classification accuracy of the closed-set samples. It can be calculated by dividing the number of correctly classified samples in $ \mathcal{Q}_{k} $ by the number of samples in $ \mathcal{Q}_{k} $. AUROC is used to measure the accuracy of open-set recognition. It is the area under the ROC curve for open-set recognition of all query samples $ \mathcal{Q}_{k} \cup \mathcal{Q}_{u} $. A larger ACC or AUROC indicates better performance.

We also calculate F1 Score, FPR95 and AUPR for the open-set recognition in our ablation study. F1 Score is the harmonic average of precision and recall.
FPR95 is short for FPR@TPR95, which is the false positive rate when the true positive rate is 95\%. The smaller the FPR95 is, the better the model is. AUPR is the area under the PR curve. Compared to AUROC on balanced datasets, AUPR is a more indicative metric in highly unbalanced datasets.

\begin{table*}[htbp]
  \centering
  \scalebox{0.9}{
  \begin{tabular}{l c c c c c c}
    \toprule
    \multirow{2}{*}{\bf Dataset} & \multirow{2}{*}{\bf Methods} & \multirow{2}{*}{\bf Publication} & \multicolumn{2}{c}{\bf 1-shot} & \multicolumn{2}{c}{\bf 5-shot} \\
    \multirow{2}{*}{} & \multirow{2}{*}{} & \multirow{2}{*}{} & ACC & AUROC & ACC & AUROC \\
    \midrule
    \multirow{12}{*}{miniImageNet} & ProtoNet~\cite{ProtoNet} & NIPS2017 & $64.01 \pm 0.88$ & $51.81 \pm 0.93$ & $80.09 \pm 0.58$ & $60.39 \pm 0.92$ \\
    \multirow{12}{*}{} & FEAT~\cite{FEAT} & CVPR2020 & $67.02 \pm 0.85$ & $57.01 \pm 0.84$ & $82.02 \pm 0.53$ & $63.18 \pm 0.78$ \\
    \multirow{12}{*}{} & PEELER~\cite{PEELER} & \multirow{2}{*}{CVPR2020} & $58.31 \pm 0.58$ & $61.66 \pm 0.62$ & $75.08 \pm 0.72$ & $69.85 \pm 0.70$ \\
    \multirow{12}{*}{} & PEELER*~\cite{PEELER} & \multirow{2}{*}{} & $65.86 \pm 0.85$ & $60.57 \pm 0.83$ & $80.61 \pm 0.59$ & $67.35 \pm 0.80$ \\
    \multirow{12}{*}{} & SnaTCHer-F~\cite{SnaTCHer} & \multirow{3}{*}{CVPR2021} & $67.02 \pm 0.85$ & $68.27 \pm 0.96$ & $82.02 \pm 0.53$ & $77.42 \pm 0.73$ \\
    \multirow{12}{*}{} & SnaTCHer-T~\cite{SnaTCHer} & \multirow{3}{*}{} & $66.60 \pm 0.80$ & $70.17 \pm 0.88$ & $81.77 \pm 0.53$ & $76.66 \pm 0.78$ \\
    \multirow{12}{*}{} & SnaTCHer-L~\cite{SnaTCHer} & \multirow{3}{*}{} & $67.60 \pm 0.83$ & $69.40 \pm 0.92$ & $82.36 \pm 0.58$ & $76.15 \pm 0.83$ \\
    \multirow{12}{*}{} & ATT~\cite{ATT} & \multirow{2}{*}{CVPR2022} & $67.64 \pm 0.81$ & $71.35 \pm 0.68$ & $82.31 \pm 0.49$ & $79.85 \pm 0.58$ \\
    \multirow{12}{*}{} & ATT-G~\cite{ATT} & \multirow{2}{*}{} & $68.11 \pm 0.81$ & $72.41 \pm 0.72$ & \bm{$83.12 \pm 0.48$} & $79.85 \pm 0.57$ \\
    \multirow{12}{*}{} & RFDNet*~\cite{RFDNet} & TMM2022 & $66.23 \pm 0.80$ & $71.37 \pm 0.80$ & $82.44 \pm 0.54$ & $80.31 \pm 0.59$ \\
    \multirow{12}{*}{} & Ours &  & \bm{$68.26 \pm 0.85$} & \bm{$73.70 \pm 0.82$} & $83.05 \pm 0.55$ & \bm{$82.29 \pm 0.60$} \\
    \midrule
    \multirow{11}{*}{tieredImageNet} & ProtoNet~\cite{ProtoNet} & NIPS2017 & $68.26 \pm 0.96$ & $60.73 \pm 0.80$ & $83.40 \pm 0.65$ & $64.96 \pm 0.83$ \\
    \multirow{11}{*}{} & FEAT~\cite{FEAT} & CVPR2020 & $70.52 \pm 0.96$ & $63.54 \pm 0.76$ & $84.74 \pm 0.69$ & $70.74 \pm 0.75$ \\
    \multirow{11}{*}{} & PEELER*~\cite{PEELER} & CVPR2020 & $69.51 \pm 0.92$ & $65.20 \pm 0.76$ & $84.10 \pm 0.66$ & $73.27 \pm 0.71$ \\
    \multirow{11}{*}{} & SnaTCHer-F~\cite{SnaTCHer} & \multirow{3}{*}{CVPR2021} & $70.52 \pm 0.96$ & $74.28 \pm 0.80$ & $84.74 \pm 0.69$ & $82.02 \pm 0.64$ \\
    \multirow{11}{*}{} & SnaTCHer-T~\cite{SnaTCHer} & \multirow{3}{*}{} & $70.45 \pm 0.95$ & $74.84 \pm 0.79$ & $84.42 \pm 0.68$ & \bm{$82.03 \pm 0.66$} \\
    \multirow{11}{*}{} & SnaTCHer-L~\cite{SnaTCHer} & \multirow{3}{*}{} & \bm{$70.85 \pm 0.99$} & $74.95 \pm 0.83$ & $85.23 \pm 0.64$ & $80.81 \pm 0.68$ \\
    \multirow{11}{*}{} & ATT~\cite{ATT} & \multirow{2}{*}{CVPR2022} & $69.34 \pm 0.95$ & $72.74 \pm 0.78$ & $83.82 \pm 0.63$ & $78.66 \pm 0.65$ \\
    \multirow{11}{*}{} & ATT-G~\cite{ATT} & \multirow{2}{*}{} & $70.58 \pm 0.93$ & $73.43 \pm 0.78$ & \bm{$85.38 \pm 0.61$} & $81.64 \pm 0.63$ \\
    \multirow{11}{*}{} & RFDNet*~\cite{RFDNet} & TMM2022 & $66.84 \pm 0.89$ & $72.68 \pm 0.76$ & $82.64 \pm 0.63$ & $80.63 \pm 0.63$ \\
    \multirow{11}{*}{} & Ours &  & $70.50 \pm 0.93$ & \bm{$75.86 \pm 0.81$} & $84.60 \pm 0.65$ & $81.95 \pm 0.72$ \\
    \midrule
    \multirow{8}{*}{CIFAR-FS} & FEAT*~\cite{FEAT} & CVPR2020 & $70.89 \pm 0.89$ & $74.83 \pm 0.79$ & $83.96 \pm 0.64$ & $82.88 \pm 0.67$ \\
    \multirow{8}{*}{} & PEELER*~\cite{PEELER} & CVPR2020 & $71.47 \pm 0.67$ & $71.28 \pm 0.57$ & $85.46 \pm 0.47$ & $75.97 \pm 0.33$ \\
    \multirow{8}{*}{} & SnaTCHer-F*~\cite{SnaTCHer} & CVPR2021 & $75.09 \pm 0.87$ & $78.15 \pm 0.77$ & $87.18 \pm 0.62$ & $85.81 \pm 0.64$ \\
    \multirow{8}{*}{} & ATT-G~\cite{ATT} & \multirow{1}{*}{CVPR2022} & $72.43 \pm 0.65$ & $76.72 \pm 0.59$ & $86.52 \pm 0.49$ & $84.64 \pm 0.38$ \\
    \multirow{8}{*}{} & RFDNet*~\cite{RFDNet} & TMM2022 & $73.83 \pm 0.92$ & $75.35 \pm 0.77$ & $85.12 \pm 0.74$ & $84.40 \pm 0.64$ \\
    \multirow{8}{*}{} & Ours &  & \bm{$76.77 \pm 0.88$} & $78.67 \pm 0.80$ & $86.74 \pm 0.66$ & $86.56 \pm 0.59$ \\
    \multirow{8}{*}{} & Ours w/o Pix &  & $76.67 \pm 0.90$ & \bm{$79.43 \pm 0.72$} & \bm{$87.63 \pm 0.62$} & \bm{$86.84 \pm 0.58$} \\
    \bottomrule
  \end{tabular}
  }
  \caption{5-way 1-shot and 5-shot results on miniImageNet, tieredImageNet and CIFAR-FS. We calculated ACC and AUROC of each model. * denotes implementation on ResNet-12. }
  \label{tab:resuts}
\end{table*}

\subsection{Implementation Details}

Following previous FSOR methods~\cite{PEELER,SnaTCHer,ATT,RFDNet}, we use ResNet-12~\cite{ResNet} as the feature extractor of our network. By default, it uses 16x drop sampling of the images and generates a feature map with 640 channel dimensions before pooling layer. We use a widely-used FSL method FEAT~\cite{FEAT} to pre-train the feature extractor. 
For the setting of hyperparameters, we set $ T $ in Eq. \ref{eq: temp} to the k value in the $ topk $ function. Because of the scale calibration module, we simply set $ M_k $ to -1 and $ M_u $ to 1 in Eq. \ref{eq: margin}. The loss weight $ \lambda $ in Eq. \ref{eq: lambda} is set to 0.1.

During the training phase, we use the SGD optimizer to train the model for over 60,000 tasks. The learning rate is set to 0.0001 for feature extractor and 0.001 for the other initially and decayed by the factor of 10 for every 12,000 tasks. We then select the model with the best results through the validation set for testing. For data augmentation, following~\cite{ATT}, in the pre-training phase, RandomCrop, ColorJitter, RandomHorizontalFilp, and RandomRotate are used; and only RandomCrop and RandomHorizontalFilp are used for meta-training and testing.

For a N-way K-shot FSOR task, we following previous FSOR methods~\cite{PEELER,SnaTCHer,ATT,RFDNet} to select 2N classes in the dataset, half of which are known classes and the other are unknown classes. For known classes, we sample K images for each class as support set and another 15 images as known query set. For each unknown classes, we only sample 15 images as unknown query set. During the testing phase, the first 75 query samples with the highest open-set scores (\ie, the energy score in Eq. \ref{eq: combine})
are taken as open-set samples to calculate the open-set recognition metrics.

\section{Results}
\subsection{Comparison Methods}

Our method GEL is compared with the FSL approach ProtoNet~\cite{ProtoNet} and FEAT~\cite{FEAT}. Since most of the existing FSOR methods are based on the metric-based FSL method, we compare the most representative method ProtoNet and the most competitive method FEAT. In order to evaluate the open-set recognition performance, we input the open-set samples into their closed-set classifier and recognize them by thresholding the entropy. We also make a comparison with existing SOTA few-shot open-set approaches PEELER~\cite{PEELER}, SnaTCHer~\cite{SnaTCHer}, ATT~\cite{ATT} and RFDNet~\cite{RFDNet}. For fair comparison, since the original PEELER and RFDNet used ResNet-18 as the feature extractor, we re-implemented a version based on ResNet-12, named PEELER* and RFDNet*. We use the source code of FEAT and SnaTCHer-F from github to do the experiment on the CIFAR-FS. In addition, SEMAN-G~\cite{ATT} uses additional word embedding information, which we do not have, and so its results cannot be fairly compared to GEL and the other methods.

\subsection{FSOR Results}

\begin{figure}
  \centering
  \subcaptionbox{SnaTCHer-F, IoU=0.57}{
    \includegraphics[width=0.47\linewidth]{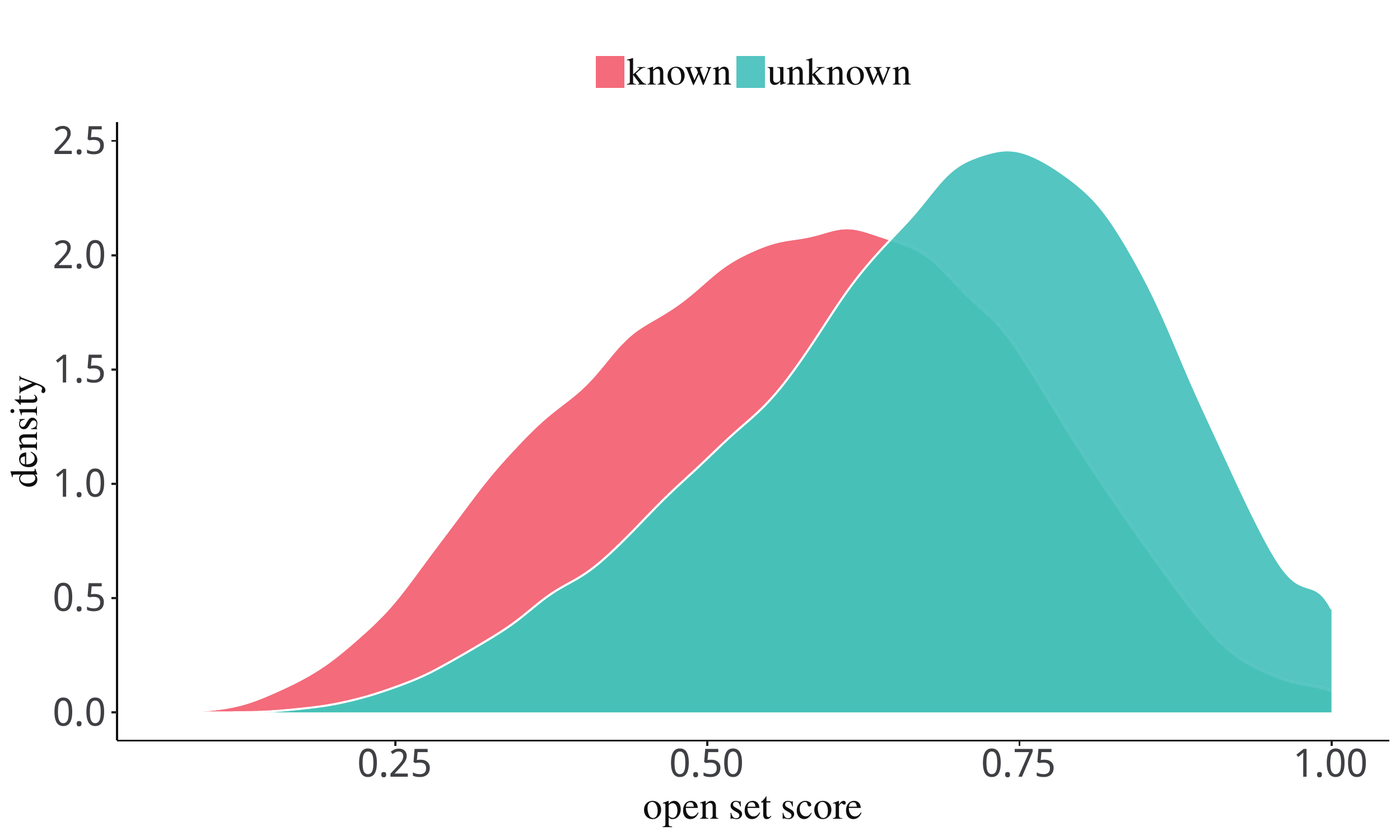}
  }
  \subcaptionbox{ATT-G, IoU=0.53}{
    \includegraphics[width=0.47\linewidth]{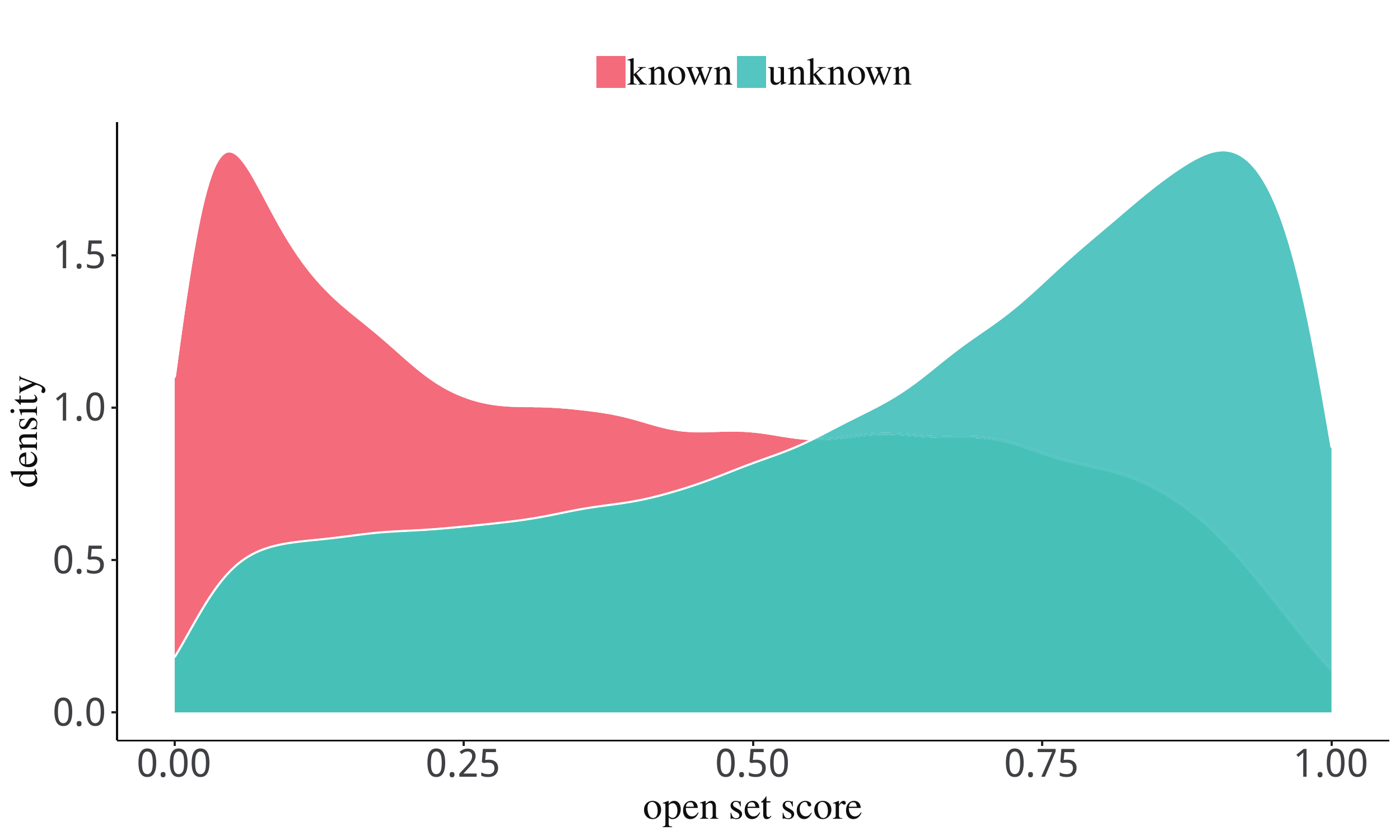}
  }
  
  \subcaptionbox{RFDNet, IoU=0.55}{
    \includegraphics[width=0.47\linewidth]{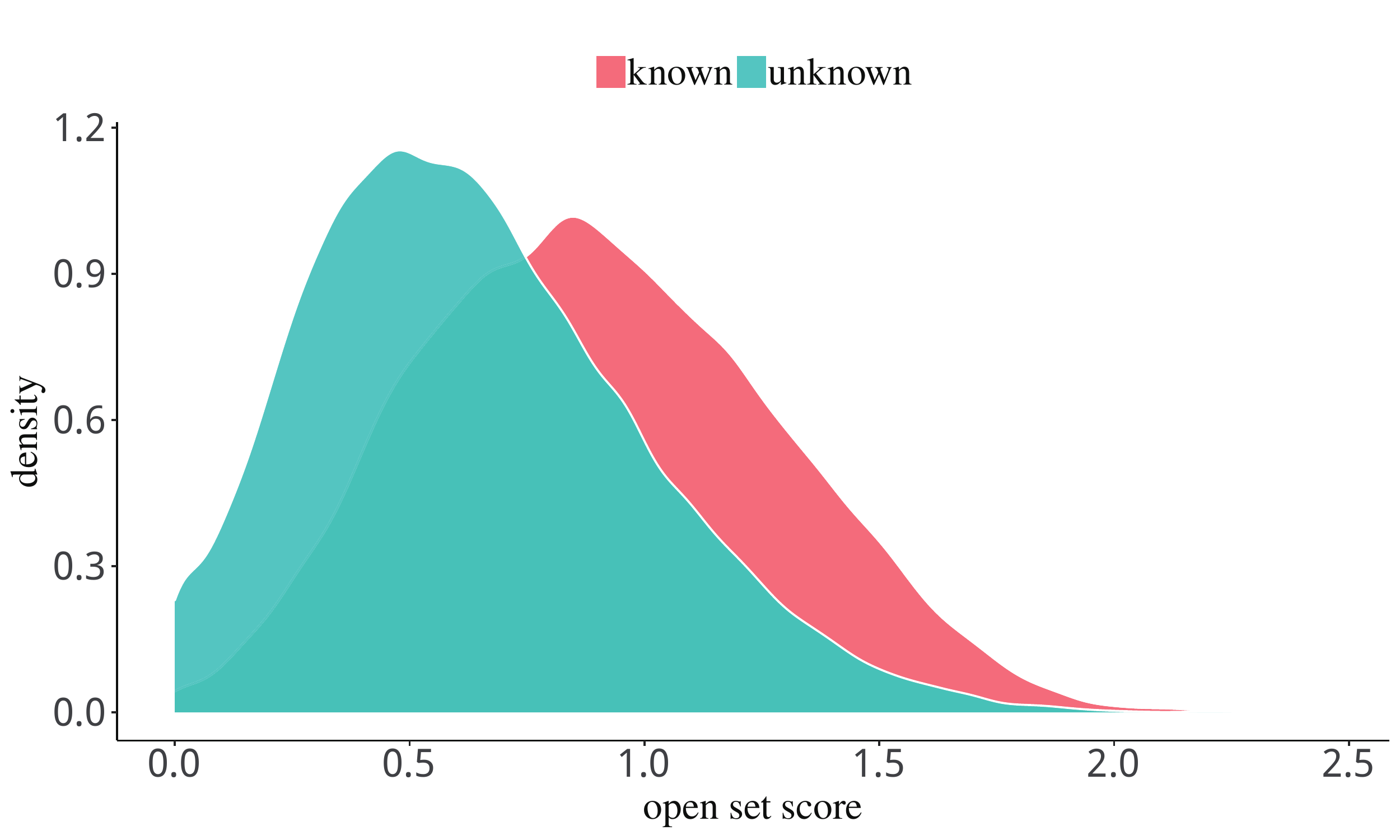}
  }
  \subcaptionbox{Ours, IoU=0.47}{
    \includegraphics[width=0.47\linewidth]{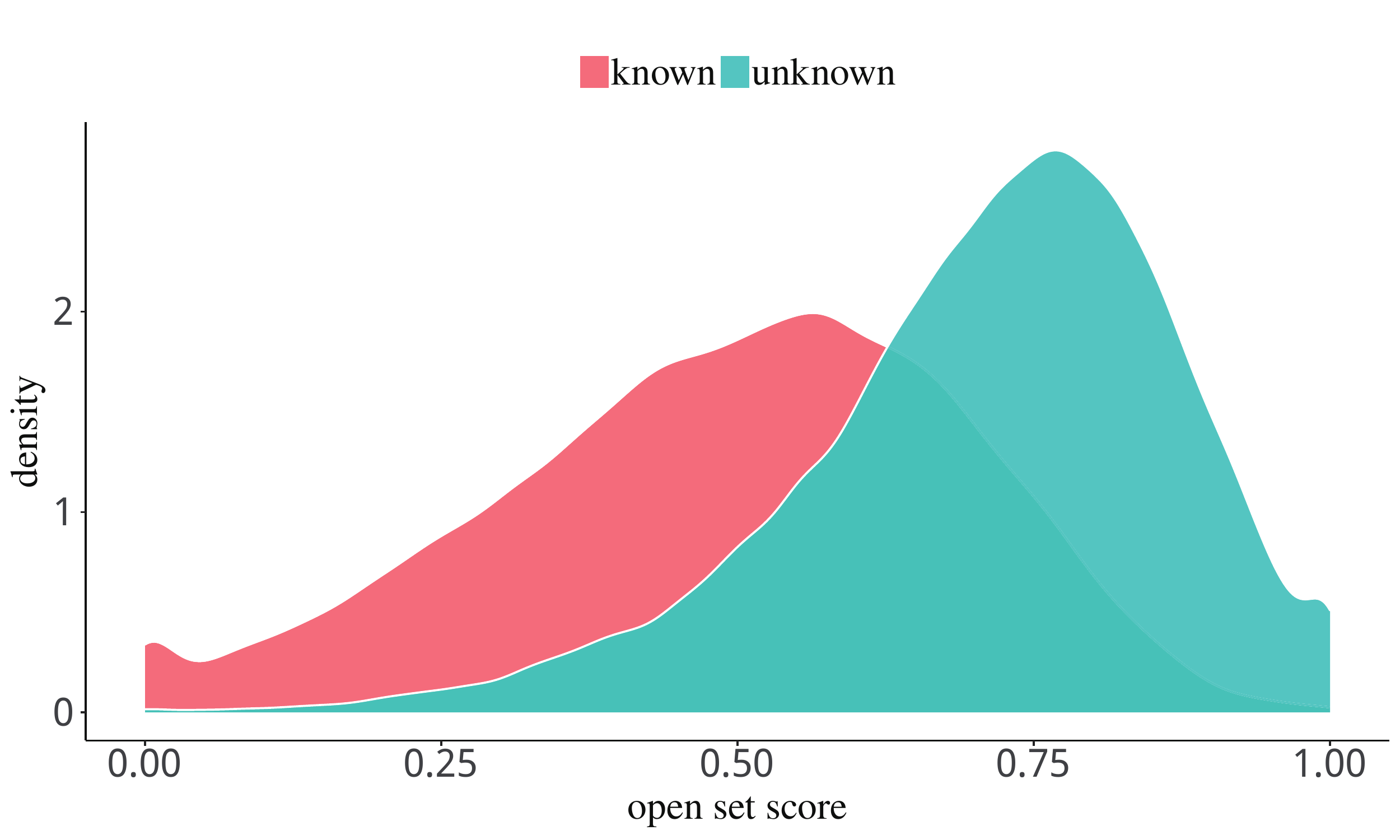}
  }
  
  \caption{Density map for normalized score histogram on miniImageNet 5-way 1-shot. By sampling 600 tasks for each method, we first make the open set score non-negative, then use the maximum value in each task for normalization. We can see that our method can better distinguish between known and unknown samples.}
  \label{fig: Normalized score histogram}
\end{figure}

Table \ref{tab:resuts} shows the results of our method GEL compared to the others on three datasets and two different shots. As can be observed, although the FSL method has a good performance in closed-set classification, its open-set recognition is poor. Compared to the previous FSOR methods, our method is substantially more effective: it obtains the best open-set recognition results (\ie, AUROC) across the three datasets in the 1-shot setting, and it is the best performer in AUROC on two datasets in the 5-shot setting, while at the same time maintaining the very competitive closed-set classification accuracy.
It is worth mentioning that compared with the other two datasets, the image resolution of CIFAR-FS is very small. After the feature extraction, the spatial dimension of the feature map is only $ 2 \times 2 $, which makes it difficult for the pixel branch to learn useful local information. Therefore, we also give the results without using the pixel branch on this dataset (\ie, \textbf{Ours w/o Pix}). It can be seen from the results that when the image size is small, only using the energy-based open-set recognizer can often achieve better results.

\begin{figure}[ht]
  \centering
  \scalebox{0.85}{
  \subcaptionbox{Ablation study of the value of $topk$ on miniImageNet under 5-way 1-shot setting.}{
    \includegraphics[width=0.88\linewidth]{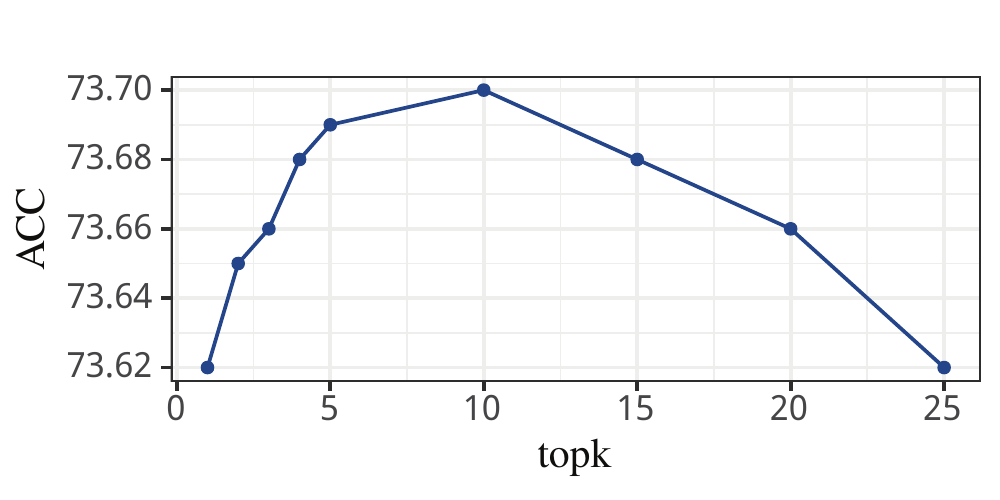}
  }
  }
  
  \scalebox{0.85}{
  \subcaptionbox{Ablation study of margin distance on miniImageNet under 5-way 1-shot setting.}{
    \includegraphics[width=0.88\linewidth]{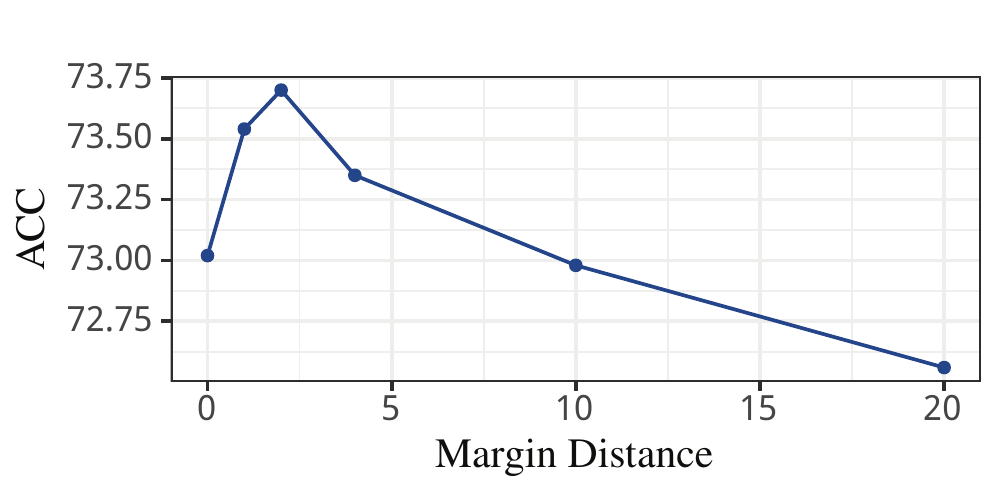}
  }
  }
  \caption{Ablation study of two hyperparameters on mimiImageNet under 5-way 1-shot setting .}
  \label{fig:k ablation}
\end{figure}

To further demonstrate the power of our model, we compare the open-set score distribution for known and unknown samples in miniImageNet 5-way 1-shot in 600 tasks. Figure \ref{fig: Normalized score histogram} compares the normalized score histogram of open-set scores between SnaTCHer-F, ATT-G, RFDNet and Our method. Since the open-set score (energy for ours and entropy for others) of known samples in our method is usually negative, we first subtract the minimum value in each task if it is negative. Then we normalized the open-set scores of all samples obtained from each task by dividing it by the maximum value, and subsequently calculate the density. We use the intersection over union (IoU) to quantitatively evaluate the overlap between the known and unknown distribution (the smaller the better), i.e. OSR capability. As can be seen from the figure, the overlap of two different color areas of our method is smaller. In other words, our method can better distinguish the known and unknown samples.

\subsection{Ablation Study}

\label{sec ablation}

Table \ref{table:ablation study} presents the ablation study results of the 5-way 1-shot experiments on two different datasets for our three proposed modules. In addition to ACC and AUROC, we also calculate F1 Score, FPR95 and AUPR of open-set recognition. From the results, it can be seen that all the three proposed modules we propose can improve the performance of open-set recognition while keeping the closed-set classification ability almost unchanged.

\begin{table*}
  \centering
  \scalebox{0.88}{
  \begin{tabular}{lcccccc}
    \toprule
    \bf{Method} & \bf{Dataset} & \bf{ACC$ \uparrow $} & \bf{AUROC$ \uparrow $} & \bf{F1 Score$ \uparrow $} & \bf{FPR95$ \downarrow $} & \bf{AUPR$ \uparrow $}\\
    \midrule
    Baseline & \multirow{4}{*}{miniImageNet} & \bm{$68.53 \pm 0.78$} & $69.85 \pm 0.87$ & $65.06 \pm 0.69$ & $79.35 \pm 1.45$ & $70.53 \pm 0.83$\\ 
    + Energy Loss & \multirow{4}{*}{} & $68.14 \pm 0.78$ & $71.97 \pm 0.83$ & $65.92 \pm 0.69$ & $77.88 \pm 1.34$ & $71.96 \pm 0.83$\\
    + Pixel wise & \multirow{4}{*}{} & $68.40 \pm 0.86$ & $72.49 \pm 0.76$ & $67.38 \pm 0.69$ & $74.99 \pm 1.35$ & $73.28 \pm 0.85$\\
    + Combine score & \multirow{4}{*}{} & $68.26 \pm 0.85$ & \bm{$73.70 \pm 0.82$} & \bm{$67.72 \pm 0.70$} & \bm{$74.10 \pm 1.38$} & \bm{$73.80 \pm 0.87$}\\
    \midrule
    Baseline & \multirow{4}{*}{tieredImageNet} & $70.55 \pm 0.93$ & $71.66 \pm 0.82$ & $67.02 \pm 0.63$ & $74.66 \pm 1.40$ & $71.01 \pm 0.83$\\
    + Energy Loss & \multirow{4}{*}{} & \bm{$70.60 \pm 0.94$} & $73.57 \pm 0.84$ & $67.90 \pm 0.61$ & $72.60 \pm 1.46$ & $73.48 \pm 0.86$\\
    + Pixel wise & \multirow{4}{*}{} & $70.39 \pm 0.96$ & $74.06 \pm 0.73$ & $67.90 \pm 0.61$ & $71.72 \pm 1.32$ & $73.10 \pm 0.75$\\
    + Combine score & \multirow{4}{*}{} & $70.50 \pm 0.93$ & \bm{$75.86 \pm 0.81$} & \bm{$69.44 \pm 0.68$} & \bm{$70.22 \pm 1.47$} & \bm{$75.78 \pm 0.80$}\\
    \bottomrule
  \end{tabular}
  }
  \caption{Ablation study of three modules proposed in GEL. We report the 5-way 1-shot results on both miniImageNet and tieredImageNet to demonstrate the effectiveness of our modules.}
  \label{table:ablation study}
\end{table*}

\begin{table}
  \centering
  \subcaptionbox{miniImageNet 5-way 1-shot}{
  \scalebox{0.75}{
  \begin{tabular}{lcc}
    \toprule
    \bf{Method} & \bf{ACC} & \bf{AUROC}\\
    \midrule
    Delay Combine & $68.39 \pm 0.86$ & $73.17 \pm 0.83$\\
    Ahead Combine & $68.26 \pm 0.85$ & $73.70 \pm 0.82$\\
    \bottomrule
  \end{tabular}
  }
    }
    \subcaptionbox{miniImageNet 5-way 5-shot}{
  \scalebox{0.75}{
    \begin{tabular}{lccc}
    \toprule
    \bf{Method} & \bf{ACC} & \bf{AUROC}\\
    \midrule
    Delay Combine & $83.00 \pm 0.55$ & $80.67 \pm 0.65$\\
    Ahead Combine & $83.05 \pm 0.55$ & $82.29 \pm 0.60$\\
    \bottomrule
  \end{tabular}
  }
  }
  \caption{Ablation study of glocal energy score combination of the class-wise branch and pixel-wise branch on miniImageNet under 5-way 1-shot and 5-shot setting.}
  \vspace{-0.5cm}
  \label{table: combine ablation}
\end{table}

In addition, we also evaluate the effect of k in Eq. \ref{eq: temp} and distance between $ M_k $ and $ M_u $ in Eq. \ref{eq: margin} on the performance. As shown in Figure \ref{fig:k ablation} (a), for the selection of different k values in our method, the variation range of AUROC is only 0.08, indicating that our method is robust to the selection of this hyperparameters. And for the choice of margin distance, in the experiment, we calculate the size of the two margins with zero as the center. As shown in Figure \ref{fig:k ablation} (b), the margin distances is greater than zero, and a smaller margin often enables the model to perform better.

For the calculation of open-set scores, in addition to combining the energy calculation as we did, called \textit{ahead combine} below, the energy can be calculated only for the pixel branch at first, and then combined with the class-wise similarity, which is denoted as \textit{delay combine}. We perform the ablation study on miniImageNet with the above two methods. As can be seen in Table \ref{table: combine ablation}, the \textit{ahead combine} method improves the performance of open-set recognition, especially when the shot is large.

\begin{table}
  \centering
  \subcaptionbox{miniImageNet 5-way 5-shot}{
  \scalebox{0.80}{
      \begin{tabular}{lcc}
        \toprule
        \bf{Method} & \bf{ACC} & \bf{AUROC}\\
        \midrule
        Fixed value & $83.05 \pm 0.55$ & $82.29 \pm 0.60$\\
        Learnable & $83.08 \pm 0.56$ & $81.37 \pm 0.64$\\
        Task-adaptive & $83.00 \pm 0.56$ & $81.34 \pm 0.62$\\
        \bottomrule
  \end{tabular}
  }
  }
 \subcaptionbox{tieredImageNet 5-way 5-shot}{
  \scalebox{0.80}{
      \begin{tabular}{lcc}
        \toprule
        \bf{Method} & \bf{ACC} & \bf{AUROC}\\
        \midrule
        Fixed value & $84.60 \pm 0.65$ & $81.95 \pm 0.72$\\
        Learnable & $84.84 \pm 0.65$ & $80.86 \pm 0.70$\\
        Task-adaptive & $84.79 \pm 0.66$ & $80.49 \pm 0.72$\\
        \bottomrule
  \end{tabular}
  }
  }
  \vspace{-0.1cm}
  \caption{Ablation study of combination coefficients between class-wise energy and pixel-wise energy on miniImageNet and tieredImageNet under 5-way 5-shot setting.}
  \vspace{-0.7cm}
  \label{table: coefficient ablation}
\end{table}

We also conduct an ablation study on the combination coefficient between class-wise energy and pixel-wise energy in Eq. \ref{eq: combine}. Three methods are used: \textit{fixed value}, \textit{learnable coefficient}, and \textit{task-adaptive} learnable coefficient on miniImageNet 5-way 5-shot. For method with a fixed value, we simply set the weight of the two branches to one. For the method with learnable coefficient, we use two learnable parameters with an initial value of one to learn the coefficients, and for the method with task-adaptive learnable coefficient, we use a linear layer to generate two coefficients from the prototypes in each task. As shown in Table \ref{table: coefficient ablation}, although learnable coefficient methods can slightly improve the classification ability of the closed-set samples, the fixed coefficient method can greatly improve the recognition ability of detecting open-set samples.

\section{Conclusion}

In this paper, we explore the under-explored problem, few-shot open-set recognition (FSOR). To have holistic detection of open-set samples, we propose a novel FSOR method, called Glocal Energy-based Learning (GEL). 
GEL improves the open-set recognition capability in few-shot settings by fusing global and local information. By combining the similarities from the class-wise and pixel-wise branches, GEL learns glocal energy scores, in which large energy scores are enforced for samples that are deviated from the few-shot examples in either the class-wise features or the pixel-wise features, and small energy scores are enforced otherwise. In doing so, GEL can detect open-set samples that are similar to the closed-set samples in either the global class level or the local feature level, while the existing methods fail to do.
Extensive experiments on multiple datasets demonstrate the effectiveness of our proposed method.

\section{Acknowledgments}
This work was supported in part by the National Natural Science Foundation of Chain under Grant 62101454, Grant 62071387, and Grant U19B2037; in part by the Fundamental Research Funds for the Central Universities; and in part by the Shenzhen Fundamental Research Program under Grant JCYJ20190806160210899. P. Wang’s participation was in part supported by Australian Research Council Discovery Projects (DP220101784). G. Pang was supported in part by the Singapore Ministry of Education
(MoE) Academic Research Fund (AcRF) Tier 1 grant (21SISSMU031).

\newpage
{\small
\bibliographystyle{ieee_fullname}
\bibliography{egbib}
}
\end{document}


\title{Supplementary Material of ``Glocal Energy-based Learning for Few-Shot Open-Set Recognition"}

\author{
Haoyu Wang\textsuperscript{1}\thanks{H. Wang, G. Pang and P. Wang contributed equally in this work.} \quad
Guansong Pang\textsuperscript{2}\footnotemark[1] \quad
Peng Wang\textsuperscript{3}\footnotemark[1] \quad
Lei Zhang\textsuperscript{1} \quad
Wei Wei\textsuperscript{1} \quad
Yanning Zhang\textsuperscript{1}\thanks{Corresponding author.} \\
\textsuperscript{1}Northwestern Polytechnical University \\
\textsuperscript{2}Singapore Management University \qquad
\textsuperscript{3}University of Wollonong\\
}
\maketitle

\section{Dataset Details}

\textbf{MiniImageNet}~\cite{miniImageNet} is designed to build a lightweight but challenging dataset. It consists of 600 RGB images per class from 100 different classes from ILSVRC-12~\cite{ImageNet}. And the classes are randomly sampled from ImageNet, which has 1000 classes. As in previous work\cite{PEELER, SnaTCHer, ATT, RFDNet}, we adopt the same split of Ravi \& Larochelle~\cite{Split}, who resized all images to $ 84 \times 84 $ and use 64 classes for training, 16 for validation and 20 for testing.

\textbf{TieredImageNet}~\cite{tieredImageNet} is also a subset of ILSVRC-12. Different from miniImageNet, tieredImageNet is not randomly sampled. It extracts 34 categories form ILSVRC-12, each of which contains $ 10\sim30 $ different classes, with a total of 608 classes. Each classes has a varying number of images, with a total of 779,165 images. Unlike miniImageNet, tieredImageNet considers ImageNet's category hierarchy. The data is divided according to categories, in which 20 categories (351 classes, 448,695 images) are used as training sets, 6 categories (97 classes, 124,261 images) are used as validation sets, and 8 categories (160 classes, 206,209 images) are used as test sets. We use the same split as ~\cite{tieredImageNet} and resized all images to $ 84 \times 84 $ as the same as previous work\cite{SnaTCHer, ATT, RFDNet}.

\textbf{CIFAR-FS}~\cite{CIFSR-FS} (CIFAR100 few shots) is randomly sampled from CIFAR100~\cite{CIFAR100} by using the same criteria with miniImageNet. We use the same split as ~\cite{CIFSR-FS}. But it is much more lightweight. It also consists of 600 RGB images per class from 100 different classes and use 64 classes for training, 16 for validation and 20 for testing. But the resolution of the image is only $ 32 \times 32 $.

\section{Method Details}

In order to illustrate our proposed method more clearly, as shown in Algorithms \ref{algorithm1} and \ref{algorithm2}, we give the PyTorch like pseudo code of our proposed pixel-wise similarity module and energy-based module, respectively.

\begin{algorithm}[t]
	\caption{Pixel-wise similarity, PyTorch-like}
	\label{algorithm1}
	\textcolor[rgb]{0, 0.7, 0.3}{\# f\_query: Tensor [2NQ, 1, $m^2$, dim/2, 1]} \\
	\textcolor[rgb]{0, 0.7, 0.3}{\# \qquad \qquad is the feature map of all query samples} \\
	\textcolor[rgb]{0, 0.7, 0.3}{\# f\_proto: Tensor [1, N, 1, dim/2, $m^2$]} \\
	\textcolor[rgb]{0, 0.7, 0.3}{\# \qquad \qquad is the feature map of all classes} \\
	\textcolor[rgb]{0, 0.7, 0.3}{\# N: N-way} \\
	\textcolor[rgb]{0, 0.7, 0.3}{\# Q: Q-query} \\
	\textcolor[rgb]{0, 0.7, 0.3}{\# m: spatial dimension of feature map} \\
	\textcolor[rgb]{0, 0.7, 0.3}{\# dim: channel dimension of feature map} \\
	\textcolor[rgb]{0, 0.7, 0.3}{\# top\_k: the hyperparameter used to select topk} \\
	\textcolor[rgb]{0, 0.7, 0.3}{\# temp: temperature coefficient hyperparameter} \\
	\textcolor[rgb]{0.7, 0.3, 0}{class} PixelSimilarity(Module):\\
	\qquad\textcolor[rgb]{0.7, 0.3, 0}{def} \_\_init\_\_(args):\\
	\qquad\qquad self.top\_k = args.top\_k \\
	\qquad\qquad self.temperature = args.temp \\
	\qquad\qquad self.cos = nn.CosineSimilarity(dim=3) \\
	\ \\
	\qquad\textcolor[rgb]{0.7, 0.3, 0}{def} forward(f\_query, f\_proto):\\
	\qquad\textcolor[rgb]{0, 0.7, 0.3}{\qquad \# cosine similarity} \\
	\qquad\qquad sim = self.cos(f\_query, f\_proto) \\
	\ \\
	\qquad\textcolor[rgb]{0, 0.7, 0.3}{\qquad \# topk} \\
	\qquad\qquad sim = sim.topk(self.topk, dim=3).values \\
	\qquad\qquad sim = sim.sum(dim=[2, 3]) \\
	\qquad\qquad sim = sim / self.temperature \\
	\qquad\qquad \textcolor[rgb]{0.7, 0.3, 0}{return} sim\\
\end{algorithm}

\begin{algorithm}[t]
	\caption{Energy-based Module, PyTorch like}
	\label{algorithm2}
	\textcolor[rgb]{0, 0.7, 0.3}{\# e\_sim: Tensor [2NQ, N] is class-wise similarity} \\
	\textcolor[rgb]{0, 0.7, 0.3}{\# f\_sim: Tensor [2NQ, N] is pixel-wise similarity} \\
	\textcolor[rgb]{0, 0.7, 0.3}{\# N: N-way} \\
	\textcolor[rgb]{0, 0.7, 0.3}{\# Q: Q-query} \\
	\textcolor[rgb]{0, 0.7, 0.3}{\# m\_k: margin for closed-set samples} \\
	\textcolor[rgb]{0, 0.7, 0.3}{\# m\_u: margin for open-set samples} \\
	\textcolor[rgb]{0.7, 0.3, 0}{class} EnergyLoss(Module):\\
	\qquad\textcolor[rgb]{0.7, 0.3, 0}{def} \_\_init\_\_(args):\\
	\qquad\qquad self.m\_k = args.m\_k \\
	\qquad\qquad self.m\_u = args.m\_u \\
	\ \\
	\qquad\textcolor[rgb]{0.7, 0.3, 0}{def} forward(e\_sim, f\_sim):\\
	\qquad\textcolor[rgb]{0, 0.7, 0.3}{\qquad \# energy score} \\
	\qquad\qquad e\_egy = -torch.logsumexp(e\_sim, dim=1) \\
	\qquad\qquad f\_egy = -torch.logsumexp(f\_sim, dim=1) \\
	\qquad\qquad energy = e\_egy + f\_egy \\
	\ \\
	\qquad\textcolor[rgb]{0, 0.7, 0.3}{\qquad \# energy loss} \\
	\qquad\qquad k\_egy, u\_egy = torch.split(energy, NQ) \\
	\qquad\qquad l\_k = pow(F.relu(k\_egy - self.m\_k), 2) \\
	\qquad\qquad l\_u = pow(F.relu(self.m\_u - u\_egy), 2) \\
	\qquad\qquad l\_energy = l\_k.mean() + l\_u.mean()\\
	\qquad\qquad \textcolor[rgb]{0.7, 0.3, 0}{return} l\_energy\\
\end{algorithm}

\section{Experiments Details}

\subsection{Detailed Task Sampling}

For a N-way K-shot Q-query FSOR task, its sampling is independent of the dataset and training or testing stage. Specifically, it will randomly select 2N classes from the dataset, half of which will be regarded as known classes and the other half as unknown classes. Q samples will be sampled for each classes as query, while additional K samples will be sampled for known classes as support.

\subsection{Detailed Ablation Study}

We provide the detailed ablation results on miniImageNet, tieredImageNet and CIFAR-FS under 5-way 1-shot and 5-shot setting.

As show in Table \ref{table:ablation study}, we give the ablation study results of three modules proposed by us on all benchmarks. On miniImageNet and tirerdImageNet dataset, our three modules can improve the performance of open-set recognition while maintaining the closed-set classification performance. On the CIFAR-FS dataset, our pixel-wise similarity branch may not work due to the decrease in original image resolution. The spatial dimension of the feature map is only $ 2 \times 2 $, the pixel-wise similarity branch is difficult to get discriminative local information. But using only the energy loss, our open-set recognition performance can significantly exceed the baseline.

We also give the results of glocal energy score combination of the class-wise branch and pixel-wise on all benchmarks. It can be seen in Table \ref{table: combine ablation} that \textit{Ahead Combine} can always improve open-set recognition performance compared to \textit{Delay Combine}.

For the ablation study of combination coefficients between class-wise energy and pixel-wise energy, we give detailed results in Table \ref{table: coefficient ablation}. Compared to \textit{Learnable} and \textit{Task-adaptive}, \textit{Fixed value} achieves better open-set recognition performance on all benchmarks.

\begin{table*}
  \centering
  \renewcommand\arraystretch{1.2}
  \scalebox{1.0}{
  \begin{tabular}{clccccc}
    \toprule
    \bf{Benchmark} & \bf{Method} & \bf{ACC$ \uparrow $} & \bf{AUROC$ \uparrow $} & \bf{F1 Score$ \uparrow $} & \bf{FPR95$ \downarrow $} & \bf{AUPR$ \uparrow $}\\
    \midrule
    \multirow{4}{*}{\makecell[c]{miniImageNet \\ 5-way 1-shot}} & Baseline & $68.53 \pm 0.78$ & $69.85 \pm 0.87$ & $65.06 \pm 0.69$ & $79.35 \pm 1.45$ & $70.53 \pm 0.83$\\ 
    \multirow{4}{*}{} & + Energy Loss & $68.14 \pm 0.78$ & $71.97 \pm 0.83$ & $65.92 \pm 0.69$ & $77.88 \pm 1.34$ & $71.96 \pm 0.83$\\
    \multirow{4}{*}{} & + Pixel wise & $68.40 \pm 0.86$ & $72.49 \pm 0.76$ & $67.38 \pm 0.69$ & $74.99 \pm 1.35$ & $73.28 \pm 0.85$\\
    \multirow{4}{*}{} & + Combine score & $68.26 \pm 0.85$ & $73.70 \pm 0.82$ & $67.72 \pm 0.70$ & $74.10 \pm 1.38$ & $73.80 \pm 0.87$\\
    \midrule
    \multirow{4}{*}{\makecell[c]{miniImageNet \\ 5-way 5-shot}} & Baseline & $83.66 \pm 0.55$ & $77.46 \pm 0.78$ & $71.36 \pm 0.68$ & $70.95 \pm 1.35$ & $78.63 \pm 0.69$\\ 
    \multirow{4}{*}{} & + Energy Loss & $82.88 \pm 0.55$ & $79.25 \pm 0.69$ & $72.24 \pm 0.63$ & $69.02 \pm 1.54$ & $79.97 \pm 0.70$\\
    \multirow{4}{*}{} & + Pixel wise & $83.57 \pm 0.53$ & $80.47 \pm 0.66$ & $73.32 \pm 0.61$ & $66.18 \pm 1.50$ & $80.98 \pm 0.66$\\
    \multirow{4}{*}{} & + Combine score & $83.05 \pm 0.55$ & $82.29 \pm 0.60$ & $74.75 \pm 0.56$ & $61.52 \pm 1.44$ & $82.40 \pm 0.63$\\
    \midrule
    \multirow{4}{*}{\makecell[c]{tieredImageNet \\ 5-way 1-shot}} & Baseline & $70.55 \pm 0.93$ & $71.66 \pm 0.82$ & $67.02 \pm 0.63$ & $74.66 \pm 1.40$ & $71.01 \pm 0.83$\\
    \multirow{4}{*}{} & + Energy Loss & $70.60 \pm 0.94$ & $73.57 \pm 0.84$ & $67.90 \pm 0.61$ & $72.60 \pm 1.46$ & $73.48 \pm 0.86$\\
    \multirow{4}{*}{} & + Pixel wise & $70.39 \pm 0.96$ & $74.06 \pm 0.73$ & $67.90 \pm 0.61$ & $71.72 \pm 1.32$ & $73.10 \pm 0.75$\\
    \multirow{4}{*}{} & + Combine score & $70.50 \pm 0.93$ & $75.86 \pm 0.81$ & $69.44 \pm 0.68$ & $70.22 \pm 1.47$ & $75.78 \pm 0.80$\\
    \midrule
    \multirow{4}{*}{\makecell[c]{tieredImageNet \\ 5-way 5-shot}} & Baseline & $85.38 \pm 0.66$ & $76.32 \pm 0.71$ & $70.42 \pm 0.62$ & $68.91 \pm 1.60$ & $76.82 \pm 0.70$\\
    \multirow{4}{*}{} & + Energy Loss & $84.24 \pm 0.69$ & $79.54 \pm 0.67$ & $72.38 \pm 0.61$ & $66.40 \pm 1.55$ & $79.86 \pm 0.66$\\
    \multirow{4}{*}{} & + Pixel wise & $85.06 \pm 0.66$ & $80.22 \pm 0.72$ & $73.08 \pm 0.61$ & $63.29 \pm 1.63$ & $80.14 \pm 0.74$\\
    \multirow{4}{*}{} & + Combine score & $84.60 \pm 0.65$ & $81.95 \pm 0.72$ & $74.39 \pm 0.63$ & $59.77 \pm 1.56$ & $82.17 \pm 0.68$\\
    \midrule
    \multirow{4}{*}{\makecell[c]{CIFAR-FS \\ 5-way 1-shot}} & Baseline & $76.85 \pm 0.88$ & $78.51 \pm 0.75$ & $71.70 \pm 0.72$ & $67.02 \pm 1.72$ & $79.11 \pm 0.80$\\
    \multirow{4}{*}{} & + Energy Loss & $76.67 \pm 0.90$ & $79.43 \pm 0.72$ & $72.44 \pm 0.63$ & $65.59 \pm 1.72$ & $79.70 \pm 0.74$\\
    \multirow{4}{*}{} & + Pixel wise & $76.60 \pm 0.87$ & $78.33 \pm 0.81$ & $71.65 \pm 0.71$ & $67.12 \pm 1.72$ & $78.73 \pm 0.80$\\
    \multirow{4}{*}{} & + Combine score & $76.77 \pm 0.88$ & $78.67 \pm 0.80$ & $71.77 \pm 0.72$ & $66.93 \pm 1.72$ & $79.51 \pm 0.81$\\
    \midrule
    \multirow{4}{*}{\makecell[c]{CIFAR-FS \\ 5-way 5-shot}} & Baseline & $87.98 \pm 0.61$ & $84.93 \pm 0.64$ & $77.12 \pm 0.59$ & $57.04 \pm 1.95$ & $84.95 \pm 0.56$\\
    \multirow{4}{*}{} & + Energy Loss & $87.63 \pm 0.62$ & $86.84 \pm 0.58$ & $79.51 \pm 0.58$ & $54.47 \pm 1.98$ & $87.81 \pm 0.52$\\
    \multirow{4}{*}{} & + Pixel wise & $86.94 \pm 0.65$ & $85.55 \pm 0.59$ & $78.10 \pm 0.59$ & $55.68 \pm 1.93$ & $86.17 \pm 0.56$\\
    \multirow{4}{*}{} & + Combine score & $86.74 \pm 0.66$ & $86.56 \pm 0.59$ & $79.15 \pm 0.59$ & $55.33 \pm 2.03$ & $87.71 \pm 0.52$\\
    \bottomrule
  \end{tabular}
  }
  \caption{Ablation study of three modules proposed in GEL. We report the 5-way 1-shot and 5-shot results on miniImageNet, tieredImageNet and CIFAR-FS to demonstrate the effectiveness of our modules.}
  \label{table:ablation study}
\end{table*}

\begin{table*}[t]
  \centering
  \renewcommand\arraystretch{1.1}
  \scalebox{1.15}{
  \begin{tabular}{llcccc}
    \toprule
    &  & \multicolumn{2}{c}{\bf{5-way 1-shot}} & \multicolumn{2}{c}{\bf{5-way 5-shot}}\\
    \bf{Dataset} & \bf{Method} & \bf{ACC} & \bf{AUROC} & \bf{ACC} & \bf{AUROC}\\
    \midrule
    \multirow{2}{*}{{miniImageNet}} & Delay Combine & $68.39 \pm 0.86$ & $73.17 \pm 0.83$ & $83.00 \pm 0.55$ & $80.67 \pm 0.65$ \\
    \multirow{2}{*}{{}} & Ahead Combine & $68.26 \pm 0.85$ & $73.70 \pm 0.82$ & $83.05 \pm 0.55$ & $82.29 \pm 0.60$ \\
    \midrule
    \multirow{2}{*}{{tieredImageNet}} & Delay Combine & $70.72 \pm 0.99$ & $74.00 \pm 0.79$ & $84.44 \pm 0.66$ & $80.57 \pm 0.69$ \\
    \multirow{2}{*}{{}} & Ahead Combine & $70.50 \pm 0.93$ & $75.86 \pm 0.81$ & $84.60 \pm 0.65$ & $81.95 \pm 0.72$ \\
    \midrule
    \multirow{2}{*}{{CIFAR-FS}} & Delay Combine & $76.58 \pm 0.86$ & $78.29 \pm 0.76$ & $87.75 \pm 0.64$ & $86.01 \pm 0.63$ \\
    \multirow{2}{*}{{}} & Ahead Combine & $76.77 \pm 0.88$ & $78.67 \pm 0.80$ & $86.74 \pm 0.66$ & $86.56 \pm 0.59$ \\
    \bottomrule
  \end{tabular}
  }
  \caption{Ablation study of glocal energy score combination of the class-wise branch and pixel-wise branch on miniImageNet, tieredImageNet and CIFAR-FS under 5-way 1-shot and 5-shot setting.}
  \label{table: combine ablation}
\end{table*}

\begin{table*}
  \centering
  \renewcommand\arraystretch{1.1}
  \scalebox{1.15}{
  \begin{tabular}{llcccc}
    \toprule
    &  & \multicolumn{2}{c}{\bf{5-way 1-shot}} & \multicolumn{2}{c}{\bf{5-way 5-shot}}\\
    \bf{Dataset} & \bf{Method} & \bf{ACC} & \bf{AUROC} & \bf{ACC} & \bf{AUROC}\\
    \midrule
    \multirow{3}{*}{{miniImageNet}} & Fixed value & $68.26 \pm 0.85$ & $73.70 \pm 0.82$ & $83.05 \pm 0.55$ & $82.29 \pm 0.60$ \\
    \multirow{3}{*}{{}} & Learnable & $68.34 \pm 0.85$ & $73.40 \pm 0.82$ & $83.08 \pm 0.56$ & $81.37 \pm 0.64$ \\
    \multirow{3}{*}{{}} & Task-adaptive & $68.34 \pm 0.85$ & $73.71 \pm 0.72$ & $83.00 \pm 0.56$ & $81.34 \pm 0.62$\\
    \midrule
    \multirow{3}{*}{{tieredImageNet}} & Fixed value & $70.50 \pm 0.93$ & $75.86 \pm 0.81$ & $84.60 \pm 0.65$ & $81.95 \pm 0.72$ \\
    \multirow{3}{*}{{}} & Learnable & $70.99 \pm 0.93$ & $75.59 \pm 0.80$ & $84.84 \pm 0.65$ & $80.86 \pm 0.70$ \\
    \multirow{3}{*}{{}} & Task-adaptive & $70.68 \pm 0.92$ & $74.93 \pm 0.83$ & $84.79 \pm 0.66$ & $80.49 \pm 0.72$ \\
    \midrule
    \multirow{3}{*}{{CIFAR-FS}} & Fixed value & $76.77 \pm 0.88$ & $78.67 \pm 0.80$ & $86.74 \pm 0.66$ & $86.56 \pm 0.59$ \\
    \multirow{3}{*}{{}} & Learnable & $76.60 \pm 0.87$ & $78.62 \pm 0.80$ & $86.94 \pm 0.65$ & $87.29 \pm 0.57$ \\
    \multirow{3}{*}{{}} & Task-adaptive & $76.70 \pm 0.87$ & $77.99 \pm 0.78$ & $87.33 \pm 0.60$ & $85.67 \pm 0.60$ \\
    \bottomrule
  \end{tabular}
  }
  \caption{Ablation study of combination coefficients between class-wise energy and pixel-wise energy on miniImageNet, tieredImageNet and CIFAR-FS under 5-way 1-shot and 5-shot setting.}
  \label{table: coefficient ablation}
\end{table*}

\begin{table*}[ht]
  \centering
  \scalebox{1.15}{
  \begin{tabular}{lcccccc}
    \toprule
    \multirow{2}{*}{\bf Dataset} & \multirow{2}{*}{\bf Method} &\multicolumn{2}{c}{\bf 5-way 1-shot} & \multicolumn{2}{c}{\bf 5-way 5-shot} \\
    \multirow{2}{*}{} & \multirow{2}{*}{} & ACC & AUROC & ACC & AUROC \\
    \midrule
    MiniImageNet & RFDNet & $66.40 \pm 0.82$ & $71.91 \pm 0.78$ & $81.91 \pm 0.60$ & \bm{$81.29 \pm 0.56$}\\
    MiniImageNet & Ours & \bm{$67.40 \pm 0.84$} & \bm{$72.35 \pm 0.80$} & \bm{$82.32 \pm 0.56$} & $80.89 \pm 0.65$\\
    \midrule
    TieredImageNet & RFDNet & $70.33 \pm 0.90$ & $75.12 \pm 0.74$ & \bm{$84.88 \pm 0.62$} & \bm{$82.19 \pm 0.66$}\\
    TieredImageNet & Ours & \bm{$70.45 \pm 0.92$} & \bm{$75.69 \pm 0.83$} & $84.21 \pm 0.67$ & $81.47 \pm 0.76$\\
    \midrule
    CIFAR-FS & RFDNet & $71.34 \pm 0.90$ & $76.74 \pm 0.75$ & $84.61 \pm 0.62$ & $85.17 \pm 0.57$\\
    CIFAR-FS & Ours & \bm{$73.36 \pm 0.89$} & \bm{$77.69 \pm 0.82$} & \bm{$85.46 \pm 0.63$} & \bm{$85.68 \pm 0.64$}\\
    \bottomrule
  \end{tabular}
  }
  \caption{Results of our method with ResNet-18 backbone. We reimplement our method using ResNet-18 as RFDNet, and report their comparison in the table. Our method still outperforms RFDNet in most cases.}
  \label{table:resnet-18}
\end{table*}

\begin{figure*}[ht]
  \centering
  \scalebox{1.35}{
    \includegraphics[width=0.71\linewidth]{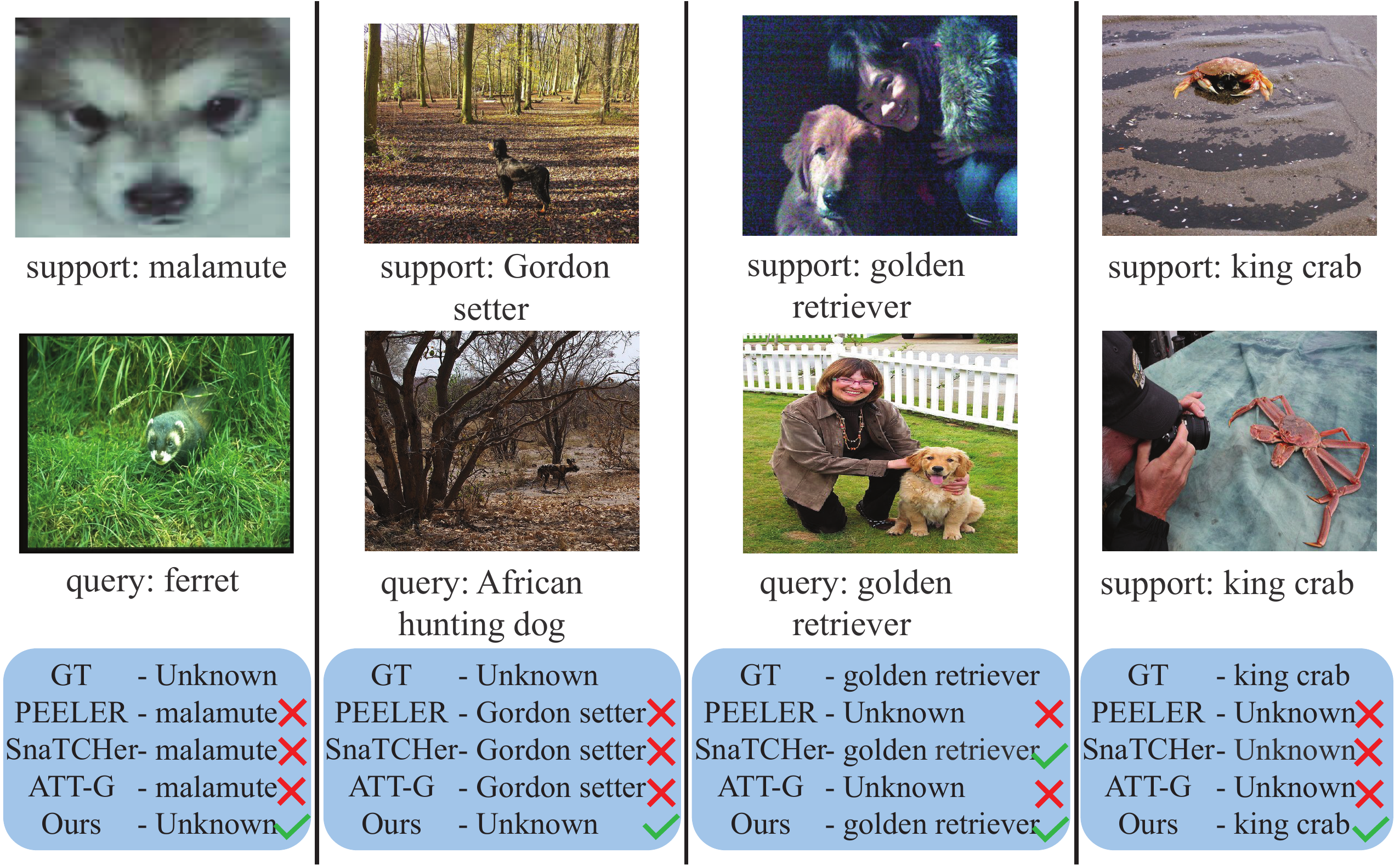}
    }
  \caption{Four additional examples.}
  \label{fig: More pixel motivation}
\end{figure*}

\newpage
{\small
\bibliographystyle{ieee_fullname}
\bibliography{egbib}
}